\documentclass[runningheads]{llncs}
\usepackage[T1]{fontenc}
\usepackage{graphicx}
\begin{document}
\title{Quantifying Cancer Likeness: A Statistical Approach for Pathological Image Diagnosis}
\titlerunning{Statistical Approach for Pathological Image Diagnosis}
%
\author{Toshiki Kindo\inst{1}\orcidID{0000-0003-0457-3706} 
}
\authorrunning{T.KINDO}
%
\institute{Kanazawa Institute of Technology\\
\email{toshiki.kindo@neptune.kanazawa-it.ac.jp}\\
 }
%
\maketitle              
\begin{abstract}
In this paper, we present a new statistical approach to automatically identify cancer regions in pathological images. The proposed method is built from statistical theory in line with evidence-based medicine. The two core technologies are the classification information of image features, which was introduced based on information theory and which cancer features take positive values, normal features take negative values, and the calculation technique for determining their spatial distribution.
This method then estimates areas where the classification information content shows a positive value as cancer areas in the pathological image. The method achieves  AUCs of 0.95 or higher in cancer classification tasks. In addition, the proposed method has the practical advantage of not requiring a precise demarcation line between cancer and normal. This frees pathologists from the monotonous and tedious work of building consensus with other pathologists.

\end{abstract}
\section{Introduction}

Active research is underway in the application of artificial intelligence to the diagnosis of pathological images, exemplified by events such as the 2016 international competition, CAMELYON16. The findings from CAMELYON16 suggest that the diagnostic capability of artificial intelligence equals or surpasses that of medical professionals. However, amidst the growing societal anticipation for AI-assisted pathological diagnosis, commonly referred to as "AI doctors," concerns have been raised regarding the lack of diagnostic basis\cite{CAMELYON16}.

In the domain of explainable AI research, efforts to empower AI doctors with the capacity to elucidate diagnoses, as demonstrated by methodologies like Grad-CAM\cite{Grad-CAM}, are gaining prominence. Within explainable AI, aligning the human gaze point with the areas of focus identified by artificial intelligence is deemed valuable. However, given the profound implications of medical diagnosis on individuals' health and even life and death, mere coincidences in gaze points cannot be deemed adequate explanations. We believe that diagnostic evidence provided by artificial intelligence should, at the very least, be supported by statistical analysis or its equivalent.

In this paper, a novel pathological image diagnosis method, whose outline is proposed at \cite{VOID-JAMIT},  is introduced, aiming to statistically estimate cancerous regions. This method is developed by refining classical statistical processing techniques, incorporating a framework capable of identifying covariant shifts\cite{CovariantShift}, along with a partial solution for their mitigation.

\section{Statistical theory of Useful Image Features}

\subsection{Kullback-Leibler Divergence of Feature Distribution on Classification Space} 

The proposed method's theoretical feature involves defining a classification space composed of binary values indicating cancerous and normal regions. It also entails examining the distribution of local image features $f$ within a pathological image in this classification space. The distribution of local image features, denoted as $\rho^p(f)$ and $\rho^n(f)$, respectively representing cancer and normality, conforms to the equation:
\begin{equation}
\rho^p(f)+\rho^n(f)=1
\end{equation}
where $p$ and $n$ indicate cancer and normality respectively.

Here, given that the probabilities in the population, denoted as $\rho^p(f)$ and $\rho^n(f)$, are established as prior knowledge, we will delve into the theory underpinning the proposed statistical method.

First, the feature $f$, which appears with equal probability in cancerous and normal regions, 
$
\rho^p(f)\sim \rho^n(f)\sim \frac{1}{2}.
$
does not serve as a clue when distinguishing between cancer and normal. The features that are useful for the above identification are those that have a biased probability distribution in the classification space as follows

\begin{equation}
\rho^p(f)\gg \frac{1}{2} \quad\mbox{ or }\quad\rho^p(f)\ll \frac{1}{2}
\end{equation}
The former refers to an image feature that is frequently present in cancerous areas and seldom found in normal areas. Therefore, this feature, representing cancerousness, is termed a positive feature and is denoted as $f^p$. The latter, representing normality, is termed a negative feature and is denoted as $f^n$. Additionally, image features that do not contribute significantly to the classification are denoted as $f^0$ whose probability is $ \rho^p(f^0)= \frac{1}{2}$.

The usefulness of image features can be quantified using Kullback-Leibler divergence, a well-known measure of information content in the statistics and the infromation theory, as follows:
\begin{equation}
D_{KL}(\rho^p(f)||\rho^p(f^0))=\rho^p(f)\log\frac{\rho^p(f)}{\rho^p(f^0)}+(1-\rho^p(f))\log\frac{1-\rho^p(f)}{1-\rho^p(f^0)}.
\end{equation}
Kullback-Leibler divergence has the following properties, 
\begin{eqnarray}
D_{KL}(\rho^p(f^p)||\rho^p(f^0))&\gg&0 \quad\mbox{for positive feature $f^p$}\\
D_{KL}(\rho^p(f^0)||\rho^p(f^0))&=&0 \quad\mbox{for neutral feature $f^0$}\\\
D_{KL}(\rho^p(f^n)||\rho^p(f^0))&\gg& 0 \quad\mbox{for negative feature $f^n$}\
\end{eqnarray}
Therefore, it serves as a useful index for distinguishing between useful image features, positive features, and negative features, as opposed to those that are not useful. However, it does not differentiate between positive features $f^p$ and negative features $f^n$.

\subsection{Spatial Distribution of Classification Information Content}

The concern raised regarding Kullback-Leibler divergence in the preceding subsection can be addressed by incorporating the classification information content proposed in Ref. \cite{hoghoge-1}  as a tool for natural language processing. This information content, represented by $C_{KL}(\rho^p(f)||\rho^p(f^0))$, is given by:
\begin{eqnarray}
C_{KL}(\rho^p(f)||\rho^p(f^0))&=&\rho^p(f)\log\frac{\rho^p(f)}{\rho^p(f^0)}-(1-\rho^p(f))\log\frac{1-\rho^p(f)}{1-\rho^p(f^0)}\\
&=&\rho^p(f)\log2\rho^p(f)-(1-\rho^p(f))\log2(1-\rho^p(f))
\end{eqnarray}
This expression essentially corresponds to the second term of Kullback-Leibler divergence, with its sign reversed.
The classification information content has the following properties, 
\begin{eqnarray}
C_{KL}(\rho^p(f^p)||\rho^p(f^0))&\gg&0 \quad\mbox{for positive feature $f^p$}\\
C_{KL}(\rho^p(f^0)||\rho^p(f^0))&=&0 \quad\mbox{for neutral feature $f^0$}\\\
C_{KL}(\rho^p(f^n)||\rho^p(f^0))&\ll& 0 \quad\mbox{for negative feature $f^n$}\
\end{eqnarray}

\begin{figure}[t]
\centering
\includegraphics[width=\textwidth, bb=0 0 3002 2026]{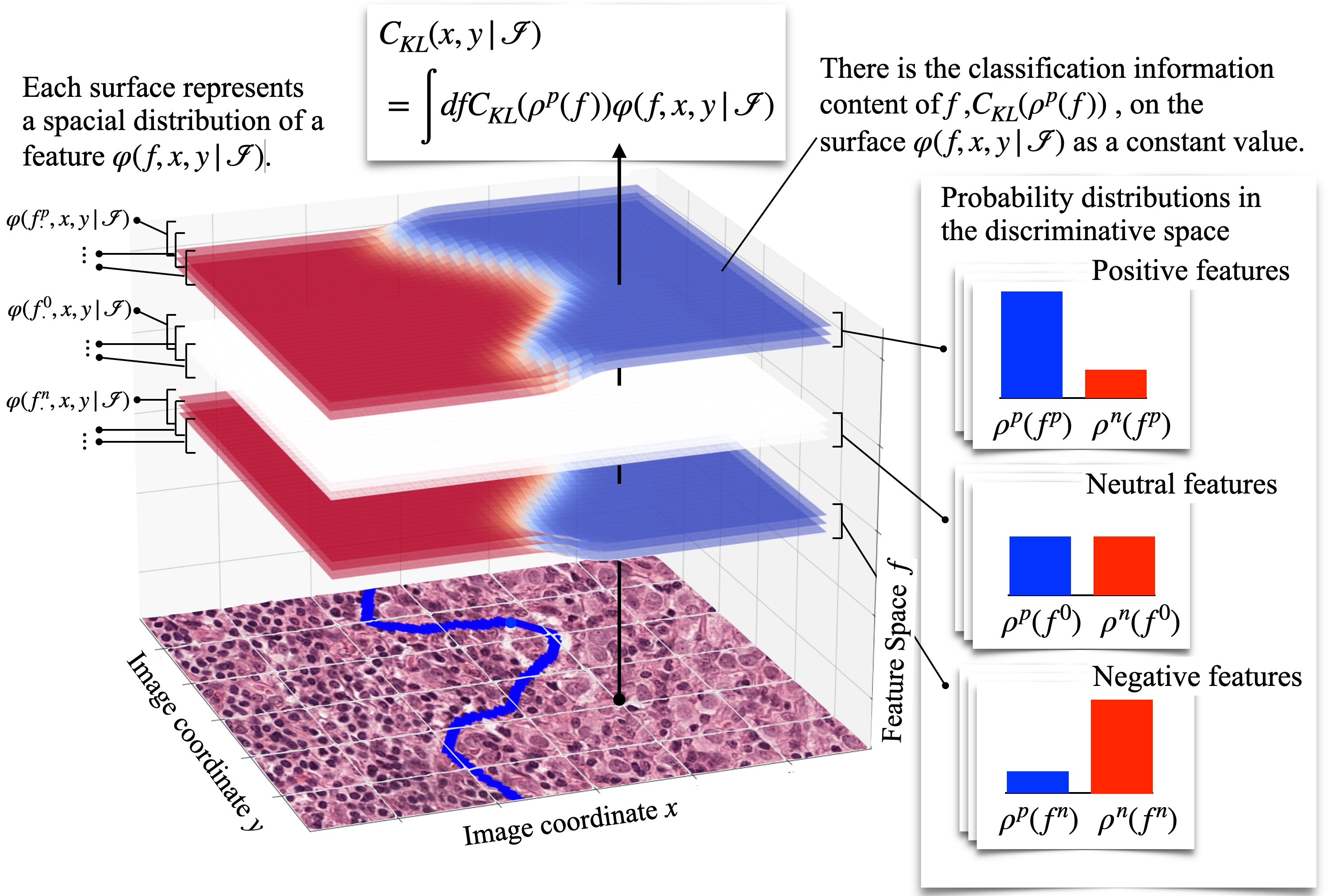}
\caption{A schematic figure shows the outline of the method proposed. The classification information content $C_{KL}(\rho^p(f)||\rho^p(f^0))$ is simply represented as $C_{KL}(\rho^p(f))$.} \label{OutlineOfVOID.jpg}
\end{figure}

To facilitate pathological image diagnosis, besides incorporating the classification information content $C_{KL}(\rho^p(f)||\rho^p(f^0))$  obtained from prior knowledge, it is imperative to consider the spatial distribution of each image feature $f$ within the pathological image under examination.
Hereinafter, $C_{KL}(\rho^p(f)||\rho^p(f^0))$ will be abbreviated as $C_{KL}(\rho^p(f))$ in this paper.

We define the spatial distribution of the image feature $f$ in the pathological image ${\cal I}$ to be diagnosed as $\varphi(f,x,y|{\cal I})$, where $(x,y)$ represents the coordinate position on the pathological image.

As a result, the spatial distribution of classification information content in the pathological image ${\cal I}$ under diagnosis is expressed by the following formula:
\begin{equation}
C_{KL}(x,y|{\cal I})=\int df C_{KL}(\rho^p(f))\varphi(f,x,y|{\cal I})\label{CKL_xy}\label{main_equation}
\end{equation}

Figure \ref{OutlineOfVOID.jpg} provides a visual representation of this concept of Equation (\ref{main_equation}).
The upward arrow indicates the integration of $C_{KL}(\rho^p(f))\varphi(f,x,y|{\cal I})$ with respect to the image feature $f$, fixing the point of image $(x, y)$. 
Then, the cancer region in the pathological image ${\cal I}$ is specified by
\begin{equation}
C_{KL}(x,y|{\cal I})>0,
\end{equation}
which represents an area where the information content indicating "This place is cancerous" exceeds the information content indicating "This place is normal."
From a statistical and information theory perspective, this forms the basis for diagnosis.

\section{Experiments using CAMELYON16 data}
When applying the above theory to actual pathological image diagnosis, it is essential to address two key challenges:
1) estimating the appearance probability of each image feature, denoted as $\rho^p(f)$, 
2) determining the spatial distribution of image features within the diagnostic target, represented as $\varphi(f,x,y|{\cal I})$.

In this section, we will offer a practical demonstration of the proposed method using data sourced from CAMELYON16, which targets the automatic detection of lymph node metastasis from breast cancer\cite{CAMELYON16}. 
We utilize SIFT with default parameters from OpenCV as the image feature\cite{SIFT}.
The parameters that we determined empirically are summarized in Table 1. The experimental results below are not sensitive to these parameters.

The original image utilized here is the highest resolution image from CAMELYON16. The pathological image measures approximately 100,000$\times$200,000 pixels, with a real-world resolution of 0.226 or 0.243 $\mu$m/pixel\cite{CAMELYON16}. Our discussion will progress using this pathological image, segmented into 191$\times$432 image patches (512$\times$512 pixels) $\{T_{XY}({\cal I})\}$, where the index $XY$ denotes the coordinates indicating the position of the two-dimensionally arranged image patches.

The spatial distribution of image features in the pathological image ${\cal I}$ to be diagnosed, $\varphi(f,x,y|{\cal I})$, is substituted by $N(f|f\in T_{XY}({\cal I}))$, representing the number of occurrences of image features equal to $f$ in an image patch $T_{XY}({\cal I})$. The equivalence between two image features is determined by the matching threshold outlined in Table \ref{ExperimentalParameters}.

Consequently, Equation  (\ref{CKL_xy}) can be reformulated as an equation for computing the classification information content of an image patch as follows:
\begin{equation}
C_{KL}(X,Y|{\cal I})=\sum_i C_{KL}(\rho^p(f_i)||\rho^p(f^0))N(f|f\in T_{XY}({\cal I})).\label{Cxypatch}
\end{equation}

\begin{table}[t]
\caption{Experimental parameters determined empirically}\label{ExperimentalParameters}
\centering
\begin{tabular}{|l|c|l|}\hline
Parameters& Value & remarks\\\hline\hline
Image feature &  SIFT\cite{SIFT}  &OpenCV default parameters are used.\\\hline
 & & When distance between two SIFT \\  
Matching threshold & 325&  descriptors is less than 325, \\  
& & they are considered equal.\\\hline  
Evidence acceptance criteria& 2&$\rho^p(f)>2 \rho^n(f) $ or $\rho^p(f)<2 \rho^n(f)$\\\hline
Lower limit of number& &Features that appear less often than \\
of occurrences& 10&the lower limit are ignored.\\\hline
Threshold of number of & &  It's introduced to skip image patches \\
descriptors in an image patch&3000 & including small number of cells.\\
to neglect it && \\\hline
\end{tabular}
\end{table}

\subsection{Probability Estimation within the Classification Space}

 In general the probability $\rho^p(f)$ on the classification space can be estimated straightforwardly as:
\begin{eqnarray}
\rho^p(f)&\sim&\rho^p(f,I^p,I^n)\\
&=&\frac{N(f|f\in I^p)}{N(f|f\in I^p)+N(f|f\in I^n)}.\label{rhoP}
\end{eqnarray}
Here $N(f|f\in I^p)$ and $N(f|f\in I^p)$ represent the frequencies of the appearance of image feature $f$ in the partial images $I^p$ labeled as cancer and $I^n$ labeled as normal, respectively.

The bottom line is that the problem at hand is essentially about selecting the appropriate partial images $I^p$ and $I^n$, which are sets of image patches:
\begin{eqnarray}
&\{I^p,I^n\}&
=
\{
\{
T^p(1),T^p(2),\cdots,T^p(k)
\},
\{
T^n(1),T^n(2),\cdots,T^n(k)
\}
\}
\end{eqnarray}
where $T^p(\cdot)$ and $T^n(\cdot)$  denote patches extracted from the cancer region and normal region, respectively.

From the selected image patches, we choose image features that can serve as evidence according to the criteria for evidence acceptance shown in Table 1. At this point, the number of positive evidence features, $n^p$, selected from the cancer region, and the number of negative evidence features, $n^n$, selected from the normal region, differ. 
The current normalization method for differences in the number of two types of evidence features is as follows:
\begin{eqnarray}
C_{KL}(X,Y|{\cal I})=(1-\alpha)&&\sum_{i=1}^{n^p} C_{KL}(\rho^p(f^p_i))N(f^p_i| f^p_i \in T^p_{XY}({\cal I}))\nonumber\\
+\alpha&&\sum_{j=1}^{n^n} C_{KL}(\rho^p(f^n_j))N(f^n_j| f^n_j \in T^n_{XY}({\cal I}))
\end{eqnarray}
where  $\alpha = n^p/(n^p+n^n)$, $f^p$ and  $f^n$ represent positive and negative evidence features respectively. 

\subsection{Rapid Learning Algorithm and Results}

\begin{figure}[t]
\centering
\includegraphics[height=6cm, bb=0 0 1786 848]{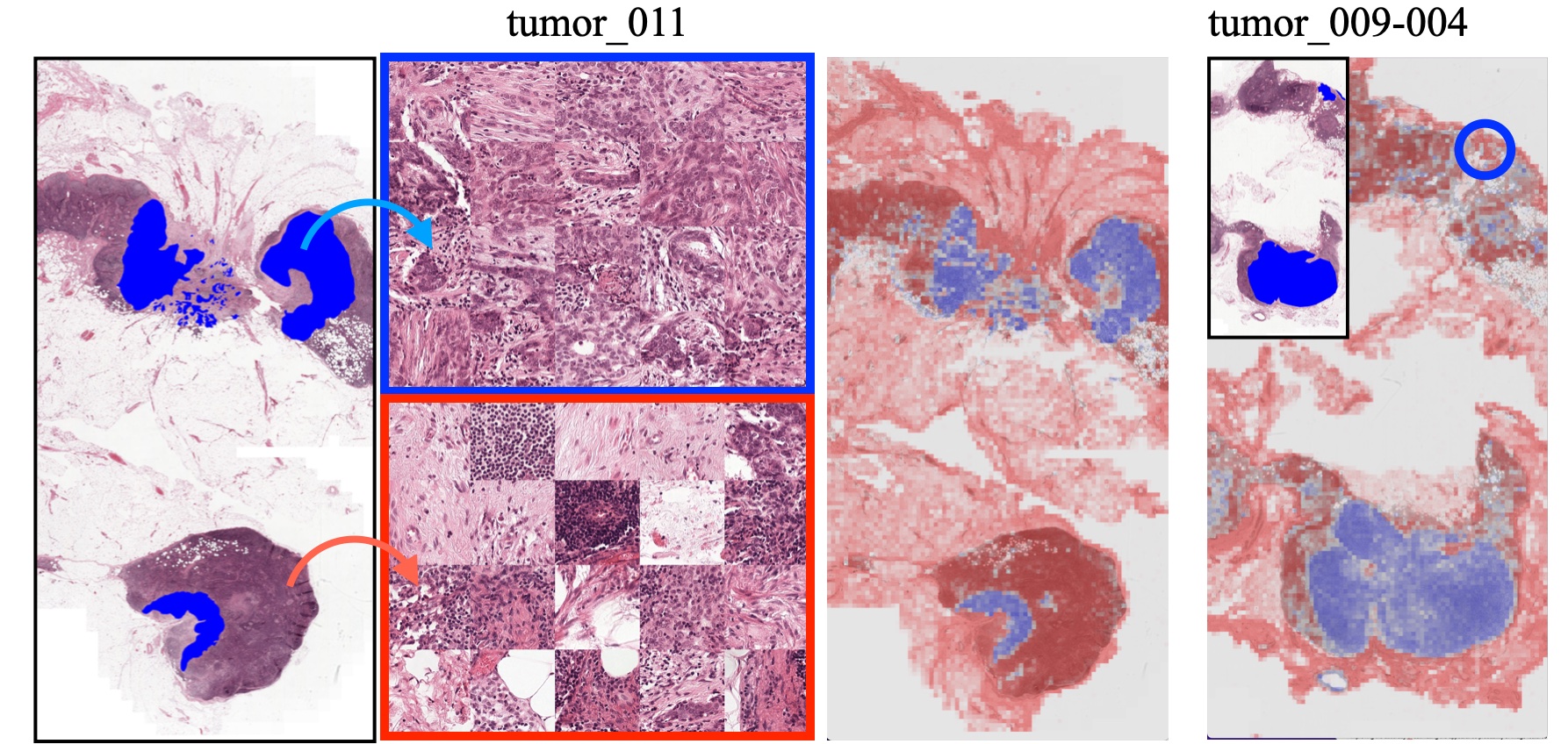}
\caption{The image on the far left displays the annotation image of tumor\_011 in the CAMELYON16 dataset.
The subsequent figure shows a total of 20 cancer patches outlined in blue and 20 normal patches outlined in red, all of which were selected from tumor\_011 in the CAMELYON16 dataset. 
The next image is the classification results for tumor\_011 are depicted, with cancerous areas marked in blue and normal areas in red. Moving on, the outcome of classifying another tumor, tumor\_009-004, using patches selected from tumor\_011 is presented with its annotation image. }\label{CICDistribution_11_1.jpg}
\end{figure}

We present a supervised learning algorithm that utilizes pathological images annotated with cancer regions. In conclusion, it's possible to detect cancerous regions by simply selecting 20 image patches from both cancerous and normal regions, as shown in Figure \ref{CICDistribution_11_1.jpg}.

If we consider selecting image patches purely from a statistical perspective, the first criterion would be to choose patches with a high density of features (high-density selection criterion). Once we have this initial selection, we can further refine it by focusing on patches with the highest negative classification information content, even if they come from cancerous regions (worst selection criterion). Additionally, there's a deterioration selection criterion, where we choose patches whose classification information content has worsened due to learning with the worst-case selection criteria. By selecting patches with deteriorated classification information content, we aim to correct any excessive deformations caused by worst-case learning.
Our current  set-up is as follows: one set consisted of learning in the order of high-density selection criteria - worst selection criteria - worse selection criteria, and after repeating this set learning three times, only learning using the worst selection criteria.

We now present typical results of our method in Figure \ref{CICDistribution_11_1.jpg}. In this figure, we display a total of 20 cancer patches outlined in blue and 20 normal patches outlined in red, all of which were selected from tumor\_011 in the CAMELYON16 dataset using our algorithm, as well as the source pathological image 'tumor\_011' and an unrelated image 'tumor\_009-004.' The blue-gray areas represent detected cancerous regions. While there is an issue with detecting cancer in a portion of the 'tumor\_009-004' image (highlighted in blue circle), overall, the results are promising considering we only used a total of 40 image patches.

\begin{figure}[t]
\centering
\includegraphics[width=12cm, bb=0 0 2484 1136]{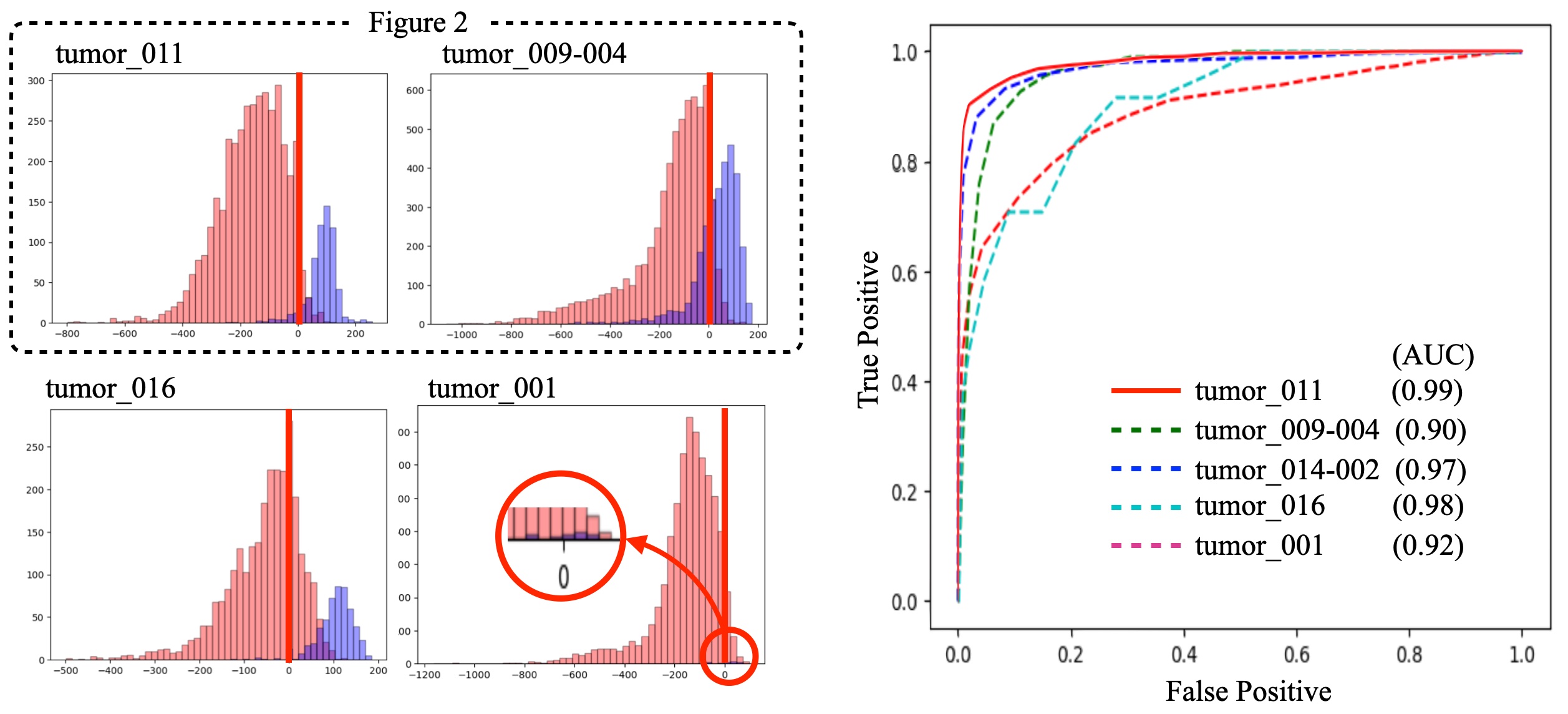}
\caption{A histogram showing the number of image patches(blue-cancer and red-normal), with the classification information content of the image patches on the horizontal axis. The vertical red line indicates the position where the classification information content is zero. } \label{Histgrams.jpg}
\end{figure}

As described above, we not only demonstrate the capability of the proposed method to detect cancerous regions but also emphasize its additional advantage: the ability to observe the distribution of the classification information content of image patches. Figure \ref{Histgrams.jpg} shows the histograms of the four images. The results shown in Figure \ref{CICDistribution_11_1.jpg} correspond to the left two histograms outlined by the dashed line. Some of the histograms with frames are from the images used for training, and the histograms to the right are from other images. By looking at this diagram, we can statistically examine the degree to which cancer and normal are separated along the classification information content. This is one of the strengths of the proposed method.

Additionally, based on the ROC curve displayed in Figure 3
, the proposed method achieved AUC of 0.95 or higher, excluding  'tomur\_001' and 'tomur\_009-004', when extracting only 40 image patches from a single pathological image. 
The issue of 'tomur\_009-004' 
is believed to stem from what's known as covariant shift\cite{CovariantShift}, signifying the inability to fully encompass the distribution within the feature space using only 40 image patches from a single pathological image. Indeed, this issue can be readily addressed by focusing on 'tumor\_009-004' and modifying the density criterion in the aforementioned algorithm to select image patches with a classification information content of 0 (no information selection criterion).

\section{Short Discussion}
Although the experimental results utilizing SIFT are presented above, the proposed method remains applicable to deep learning models employing feature acquisition functions. In a deep learning model, the output of a neuron in the pooling layer signifies the presence or absence of a specific feature within the receptive field of that neuron. Hence, the presence or absence of SIFT features in the classical method mentioned above can be substituted with this output. Consequently, by assessing the bias in the classification space of the neuron output from the pooling layer along the proposed method, it becomes feasible to determine the classification information content of the pathological image undergoing processing by the deep learning model.

Finally, I would like to emphasize that the proposed method has the practical advantage of not requiring a precise dividing line between cancer and normal. This frees pathologists from monotonous, tedious work that is difficult to reach consensus with other pathologists.

\section{Concluding Remark} 
We have proposed a statistical method for evaluating cancer likeness in pathological image diagnosis.

The proposed method represents a robust solution, embodied in a straightforward process: tallying occurrences of image features, converting them into probabilities, calculating classification information content, and aggregating them for each image patch within the pathological image. Despite its simplicity, this method offers several advantages: strong theoretical foundations in line with evidence-based medicine, good performance capable of achieving Areas Under the Curve (AUCs) of 0.90 or higher, streamlined result analysis, and liberation of pathologists from the burdensome task of annotating image creation.


\newpage

\begin{thebibliography}{8}
\bibitem{CAMELYON16}
 Ehteshami Bejnordi B, Veta M, Johannes van Diest P, van Ginneken B, Karssemeijer N, Litjens G, van der Laak JAWM, and the CAMELYON16 Consortium. Diagnostic Assessment of Deep Learning Algorithms for Detection of Lymph Node Metastases in Women With Breast Cancer. JAMA. 2017;318(22), pp. 2199--2210. doi:10.1001/jama.2017.14585 (2017).

\bibitem{Grad-CAM}
Ramprasaath R. Selvaraju, Michael Cogswell, Abhishek Das, Ramakrishna Vedantam, Devi Parikh, Dhruv Batra; Grad-CAM: Visual Explanations From Deep Networks via Gradient-Based Localization,Proceedings of the IEEE International Conference on Computer Vision (ICCV), pp. 618--626, (2017) .

\bibitem{VOID-JAMIT}
Toshiki KINDO, Shunya MUTSUDA, Sohsuke YAMADA;
Information Density Method to Evaluate the Cancer-Likeness and Normality-Likeness of Pathological Images with the Amount of Information,
MEDICAL IMAGING TECHNOLOGY 40(5), pp. 218-225,  (2022) {\it in Japanese}.

\bibitem{CovariantShift}
Hidetoshi Shimodaira,
Improving predictive inference under covariate shift by weighting the log-likelihood function,
Journal of Statistical Planning and Inference,
90,(2),
pp 227-244,
https://doi.org/10.1016/S0378-3758(00)00115-4,(2000).

\bibitem{hoghoge-1}
Toshiki Kindo, Hideyuki Yoshida, Tetsuro Morimoto, Taisuke Watanabe, Adaptive Information Filtering System that organizes personal profiles automatically, Proceedings of Joint Conference on Artificial Intelligence, 716-721, (1997).
\bibitem{SIFT}
D. G. Lowe, "Object recognition from local scale-invariant features," Proceedings of the Seventh IEEE International Conference on Computer Vision, Kerkyra, Greece, pp. 1150-1157 vol.2, doi: 10.1109/ICCV.1999.790410, (1999).

\end{thebibliography}
\end{document}